%% file: main.tex
\documentclass[fleqn,10pt]{wlscirep}
\usepackage[utf8]{inputenc}
\usepackage[T1]{fontenc}
\usepackage{soul}

\title{iSight: Towards expert–AI co-assessment for improved immunohistochemistry staining interpretation}
\author[1,*]{Jacob S. Leiby} 
\author[1,2,*]{Jialu Yao} 
\author[3,*]{Pan Lu} 
\author[3]{George Hu} 
\author[1]{Anna Davidian} 
\author[1]{Shunsuke Koga} 
\author[1]{Olivia Leung} 
\author[1]{Pravin Patel} 
\author[1,4]{Isabella Tondi Resta} 
\author[5]{Rebecca Rojansky} 
\author[1]{Derek Sung} 
\author[5]{Eric Yang} 
\author[1]{Paul J. Zhang} 
\author[5,6]{Emma Lundberg} 
\author[7]{Dokyoon Kim} 
\author[3]{Serena Yeung-Levy} 
\author[3,$\dagger$]{James Zou} 
\author[5,$\dagger$]{Thomas Montine} 
\author[8,$\dagger$]{Jeffrey Nirschl} 
\author[1,7,$\dagger$]{Zhi Huang}

\affil[1]{Department of Pathology and Laboratory Medicine, Perelman School of Medicine, University of Pennsylvania, PA, USA}
\affil[2]{Department of Electrical and Systems Engineering, University of Pennsylvania, PA, USA}
\affil[3]{Department of Biomedical Data Science, Stanford University School of Medicine, Stanford, CA, USA}
\affil[4]{Department of Pathology, University of Maryland School of Medicine, University of Maryland, MD, USA}
\affil[5]{Department of Pathology, Stanford University School of Medicine, Stanford, CA, USA}
\affil[6]{Department of Bioengineering, Stanford University School of Medicine, Stanford, CA, USA}
\affil[7]{Department of Biostatistics, Epidemiology and Informatics, Perelman School of Medicine, University of Pennsylvania, PA, USA}
\affil[8]{Department of Pathology and Laboratory Medicine, School of Medicine and Public Health, University of Wisconsin, WI, USA}

\affil[*]{Equal contribution
}
\affil[$\dagger$]{To whom the correspondence should be addressed:\newline Zhi Huang (\href{zhi.huang@pennmedicine.upenn.edu}{\textcolor{blue}{zhi.huang@pennmedicine.upenn.edu}});
Jeffrey Nirschl (\href{jjnirschl@wisc.edu}{\textcolor{blue}{jjnirschl@wisc.edu}});
Thomas Montine (\href{tmontine@stanford.edu}{\textcolor{blue}{tmontine@stanford.edu}});
James Zou (\href{jamesz@stanford.edu}{\textcolor{blue}{jamesz@stanford.edu}})
}

\begin{abstract}

Immunohistochemistry (IHC) provides information on protein expression in tissue sections and is commonly used to support pathology diagnosis and disease triage. While AI models for H\&E-stained slides show promise, their applicability to IHC is limited due to domain-specific variations. Here we introduce HPA10M, a dataset that contains 10,495,672 IHC images from the Human Protein Atlas with comprehensive metadata included, and encompasses 45 normal tissue types and 20 major cancer types. Based on HPA10M, we trained iSight, a multi-task learning framework for automated IHC staining assessment. iSight combines visual features from whole-slide images with tissue metadata through a token-level attention mechanism, simultaneously predicting staining intensity, location, quantity, tissue type, and malignancy status. On held-out data, iSight achieved 85.5\% accuracy for location, 76.6\% for intensity, and 75.7\% for quantity, outperforming fine-tuned foundation models (PLIP, CONCH) by 2.5--10.2\%. In addition, iSight demonstrates well-calibrated predictions with expected calibration errors of 0.0150-0.0408. Furthermore, in a user study with eight pathologists evaluating 200 images from two datasets, iSight outperformed initial pathologist assessments on the held-out HPA dataset (79\% vs 68\% for location, 70\% vs 57\% for intensity, 68\% vs 52\% for quantity). Inter-pathologist agreement also improved after AI assistance in both held-out HPA (Cohen's $\kappa$ increased from 0.63 to 0.70) and Stanford TMAD datasets (from 0.74 to 0.76), suggesting expert--AI co-assessment can improve IHC interpretation. This work establishes a foundation for AI systems that can improve IHC diagnostic accuracy and highlights the potential for integrating iSight into clinical workflows to enhance the consistency and reliability of IHC assessment.

\end{abstract}

\begin{document}

\flushbottom

\renewcommand{\rmdefault}{phv}
\renewcommand{\sfdefault}{phv}
\renewcommand{\familydefault}{\sfdefault}
\fontsize{11}{13.2}\selectfont
\normalfont
\setlength{\parskip}{6pt}

\maketitle
\thispagestyle{empty}

\fancyhf{}
\rfoot{\small\sffamily\bfseries\thepage}
\renewcommand{\headrulewidth}{0pt}
\renewcommand{\footrulewidth}{0pt}

\section*{Introduction}

Immunohistochemistry (IHC) is a widely used technique in diagnostic pathology and biomedical research that leverages antibody–antigen interactions to visualize specific proteins within tissue sections, providing spatially resolved molecular information. IHC plays a central role in the immunophenotypic characterization of tumors and is routinely used in clinical diagnosis and disease classification worldwide, including in the context of World Health Organization tumor classifications \cite{mebratie2024review}. Additionally, IHC serves as an important theranostic and prognostic biomarker across many tumor types, enabling patient stratification for targeted and immunotherapies, such as ER/PR and HER2 in breast cancer, HER2 in selected subsets of gastric and colorectal cancers, and PD-L1 in melanoma \cite{sheffield2016immunohistochemistry}. In translational research, IHC facilitates tumor phenotyping, evaluating assessments in animal models, and biomarker validation in clinical trials \cite{de2010immunohistochemistry}. The technique's ability to preserve tissue architecture while revealing molecular signatures makes it uniquely valuable for understanding disease mechanisms and guiding treatment decisions.

The increasing use of diagnostic and prognostic IHC testing has led to a consistent rise in IHC usage over the past two decades. At Stanford Healthcare, we observed a steady year-over-year increase in monthly IHC case volume from 2005 to 2025, with cases rising from approximately 700 per month to over 3,500 per month (\textbf{Supplementary Figure \ref{Supple Figure 1}}). This nearly five-fold increase reflects the expanding role of IHC in clinical practice. Contributing factors include expanded therapeutic indications, increasing numbers of FDA-approved antibody tests, and evolving practice patterns among younger pathologists trained in high-throughput, multiplexed IHC environments.
Despite its critical role, IHC interpretation remains labor-intensive, subjective, and inconsistently reproducible across institutions and pathologists. 
The interpretation process requires pathologists to integrate multiple variables, including staining intensity, subcellular localization, proportion of positive cells, and tissue context. As a result, IHC assessment heavily relies on expert, spatially informed visual evaluation of tissue morphology.

This landscape highlights a pressing need for scalable, automated tools to support IHC interpretation in a better and efficient way.
Such tools serve as second reads in clinical trials, support non-specialists in research settings, and facilitate quality control in clinical laboratories. Furthermore, automated IHC assessment systems could help address the growing workforce shortage in pathology, particularly in resource-limited settings and developing countries where access to subspecialty expertise is constrained. 
While recent AI models trained on hematoxylin and eosin (H\&E)-stained pathology slides have shown promise\cite{huang2023visual,ICLR2025_ebf8764e,lu2024visual,xiang2025vision,chen2024towards, ding2025multimodal}, their generalizability to IHC remains limited due to several domain-specific challenges. First, IHC exhibits substantial variations in stain color profiles depending on the chromogen used, the presence of counterstains, and differences in staining protocols across institutions. Second, unlike H\&E slides where cytologic and tissue morphology is the primary diagnostic feature, IHC interpretation requires simultaneous assessment of both morphological context and stain localization patterns. Third, IHC datasets often include rich structured metadata, such as expected protein localization, tissue type, and diagnostic codes, which are not typically available for H\&E images but could significantly enhance model performance if properly integrated. 

To address this gap, we introduce HPA10M, a large-scale, richly annotated dataset comprising 10,495,672 IHC images. Sourced from the Human Protein Atlas (HPA) \cite{uhlen2015tissue, thul2018human,thul2017subcellular}, HPA10M includes extensive metadata, such as patient demographics (e.g., age, sex), tissue or tumor metadata, SNOMED-encoded diagnosis, and if available linked Kaplan-Meier survival data, facilitating diverse use cases in pathology. 
Each image is connected to a unique MD5 hash, ensuring traceability and reproducibility, and is annotated with standardized ontologies (e.g., target protein Ensembl and UniProt, Uberon anatomy, SNOMED). The dataset encompasses 45 normal tissue types and 20 major cancer types, providing comprehensive coverage of the tissue diversity encountered in clinical practice and enabling systematic evaluation of IHC-specific staining patterns across different biological contexts.

Leveraging this dataset, we developed iSight: \textbf{\underline{I}}mmunohistochemical \textbf{\underline{S}}taining \textbf{\underline{I}}nsi\textbf{\underline{g}}ht with \textbf{\underline{H}}ybrid \textbf{\underline{T}}ransformers, a multi-task learning framework specifically designed for automated IHC assessment. Unlike previous approaches that treat IHC scoring as isolated classification problems, iSight simultaneously predicts staining intensity, subcellular localization, staining quantity, tissue type, and malignancy status, allowing the model to learn shared representations that capture both staining characteristics and tissue context. The architecture combines vision transformers for fine-grained image feature extraction with a token-level attention mechanism that aggregates information across tissue regions, which enables the model to focus on diagnostically relevant areas while maintaining global tissue context. Importantly, iSight incorporates structured metadata through a context encoding module that processes antibody information, tissue type, and diagnostic codes, effectively creating a multimodal learning framework that mirrors the way pathologists integrate clinical information during slide review. In our study, iSight achieved accuracies of 76.6\%, 85.5\%, and 75.7\% for staining intensity, subcellular localization, and staining quantity, respectively, outperforming all fine-tuned pathology foundation models adapted from H\&E analysis by 2.5–10.2\%. The model also produced well-calibrated predictions, with expected calibration errors ranging from 0.0150 to 0.0408. In a user study involving eight pathologists evaluating 200 images from two datasets, iSight exceeded initial pathologist performance on labeled HPA samples (79\% vs 68\% for localization, 70\% vs 57\% for intensity, and 68\% vs 52\% for quantity) and improved inter-pathologist agreement after AI assistance (Cohen's $\kappa$ increased from 0.63 to 0.70 on HPA and from 0.74 to 0.76 on Stanford TMAD). These expert--AI co-assessment results indicate that iSight captures clinically meaningful IHC staining patterns and supports consistent, collaborative IHC assessment.

In summary, our dataset and modeling effort supports the development of robust, foundation-scale pathology models tailored to IHC, complementing existing H\&E-based models \cite{huang2023visual,lu2024visual} and advancing the paradigm of human-AI collaboration \cite{huang2025pathologist} in clinical and research pathology. iSight represents a significant step toward clinically deployable AI systems that can augment rather than replace pathologist expertise. The framework's modular design also allows for easy adaptation to new biomarkers, scoring systems, and clinical workflows, making it a flexible platform for future development and validation studies. The curated HPA10M dataset is publicly available at \href{https://huggingface.co/datasets/nirschl-lab/hpa10m}{https://huggingface.co/datasets/nirschl-lab/hpa10m}, and the iSight model and source code are available at \href{https://github.com/zhihuanglab/iSight}{https://github.com/zhihuanglab/iSight}.

\section*{Results}

\begin{figure}
    \centering
    \includegraphics[width=1.0\linewidth]{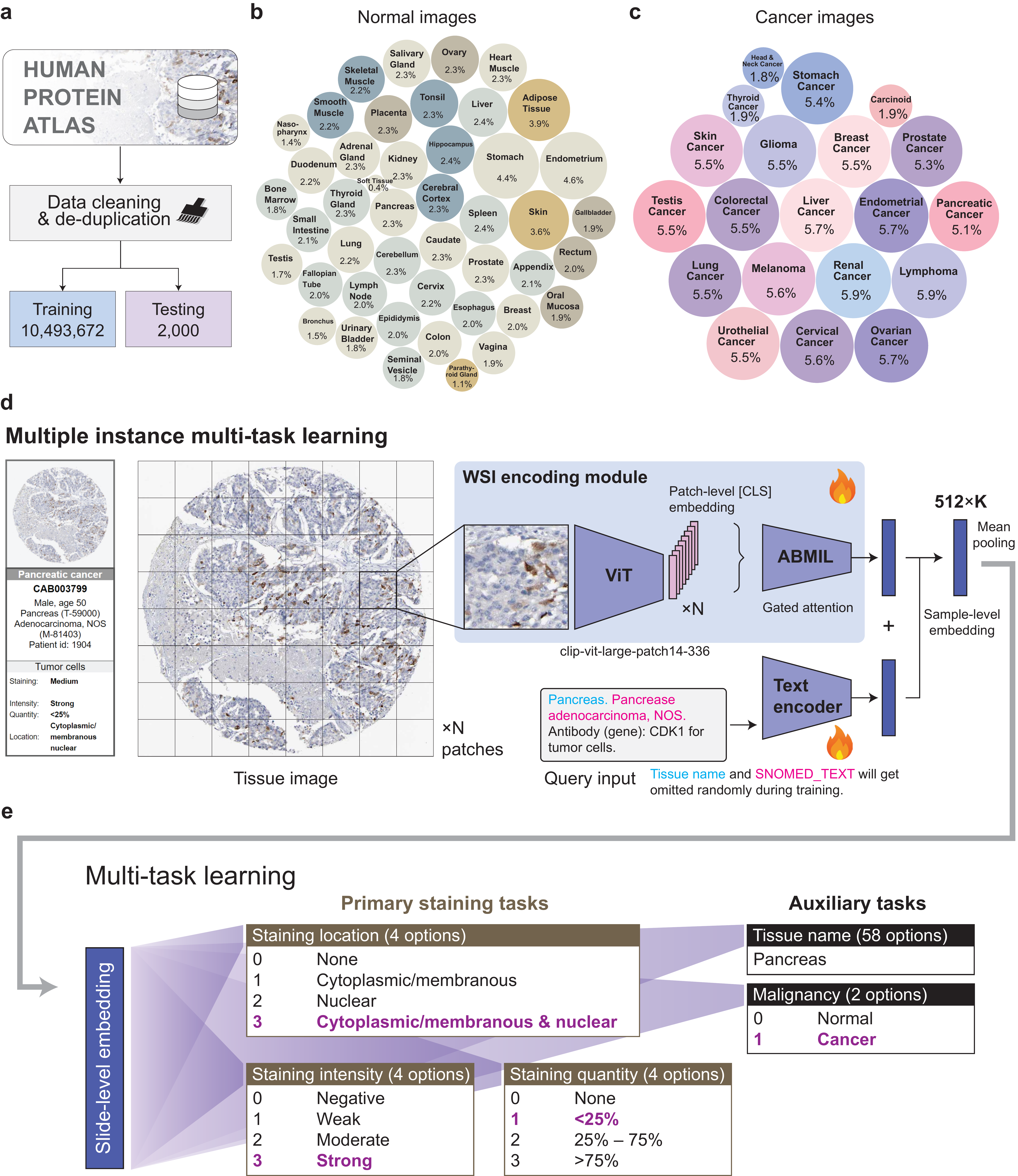}
    \caption{
    \textbf{HPA10M dataset and iSight model architecture for multi-task immunohistochemistry analysis.} 
     \textbf{a}, Dataset construction workflow from the Human Protein Atlas. 
     \textbf{b}, Distribution of 45 normal tissue types in HPA. 
     \textbf{c}, Distribution of 20 major cancer types in HPA10M.
    \textbf{d}, The model processes whole slide images by dividing them into 336×336 patches, extracting visual features with a Vision Transformer (CLIP-ViT-large-patch-14-336), and aggregating patch-level representations using gated attention-based multiple instance learning (MIL). Text metadata, including tissue type, SNOMED diagnosis, and antibody information, is encoded separately using the CLIP text encoder.
    \textbf{e},The multi-task learning framework uses five parallel classification heads to simultaneously predict staining properties (location, intensity, quantity) and tissue characteristics (tissue type, malignancy status).  Image credit: Human Protein Atlas; (\href{http://v23.proteinatlas.org/ENSG00000170312-CDK1/}{v23.proteinatlas.org/ENSG00000170312-CDK1/})
}
    \label{Figure 1}
\end{figure}

\subsection*{Establishing the large-scale IHC dataset}
We constructed HPA10M, a large-scale IHC dataset comprising over 10 million images derived from the Human Protein Atlas (HPA) version 23.0 (\textbf{Figure \ref{Figure 1}a}). The dataset provides comprehensive coverage across multiple dimensions critical for IHC assessment.
Although the source images originate from the publicly accessible HPA website \footnote{https://www.proteinatlas.org/}, these resources existed in an unstructured web format unsuitable for machine learning applications. The HPA website, developed over two decades, hosts images and protein information in an web format for easy browsing but it lacks the structure and formatting required for efficient large-scale machine learning tasks. In addition, there are no formal classification labels, harmonized metadata, or existing datasets that represent the full HPA corpora. HPA10M represents the first comprehensive curation of these web-based resources into a structured dataset with harmonized annotations, standardized metadata, and optimized formatting for machine learning tasks.

The dataset construction involved extensive curation beyond simple data collection. Images were systematically downloaded through custom web scraping, as no bulk download mechanism or API existed. The native XML metadata from the HPA website exhibited significant structural variations accumulated over 20 years of website development, requiring substantial processing to harmonize into consistent formats suitable for machine learning tasks. After the data downloaded, our pathologists performed manual and semi-automated quality checks throughout the curation process, correcting metadata errors including spelling mistakes, typos, and data entry inconsistencies.

Beyond error correction, HPA10M provides substantially enriched annotations compared to the original HPA website content. A bioinformatician and board-certified pathologist led dataset construction following a rigorous multi-stage pipeline involving standardization with biomedical ontologies (Uberon for anatomy, SNOMED CT for diagnoses), systematic error correction, redundancy management, and quality filtering. We generated novel descriptive text captions for IHC staining patterns using pathology domain expertise. Only images with complete, high-confidence metadata were retained. Each image was assigned a unique MD5 cryptographic hash, which enables identification and removal of duplicate entries (e.g., multiple protein isoforms referencing identical images). Additionally, we segmented the tissue foreground mask and bounding box for easy cropping. All metadata and MD5 hashes are traceability to the original HPA identifiers and original URL, supporting reproducibility and enabling seamless integration with additional clinical and molecular data.

HPA10M encompasses both normal and malignant tissues with comprehensive representation. The final dataset comprises 10,495,672 images, with 7,970,595 images (75.9\%) from cancer samples and 2,525,077 images  (24.1\%) from normal tissue samples. Normal tissue samples span 45 tissue types (\textbf{Figure \ref{Figure 1}b}), with major representation including endometrium (4.6\%), stomach (4.4\%), adipose tissue (3.9\%), and skin (3.6\%), alongside diverse organs with relatively balanced representation at approximately 2.0-2.4\% (including liver, spleen, hippocampus, cerebral cortex and others), and less represented tissues ranging from soft tissue (0.4\%) to bone marrow (1.8\%). Cancer samples include 20 major tumor types (\textbf{Figure \ref{Figure 1}c}), with distribution ranging from 1.8\% (head and neck cancer) to 5.9\% (lymphoma, renal cancer), including melanoma (5.6\%), cervical cancer (5.6\%), liver cancer (5.7\%), ovarian cancer (5.7\%), and endometrial cancer (5.7\%). 
Each image is associated with comprehensive metadata including tissue source, SNOMED CT-coded diagnosis, patient demographics (age, sex), molecular targets indexed via UniProt and ENSEMBL identifiers, and staining characteristics (intensity, location, quantity). In addition, the dataset encompasses over 17,200 different protein biomarkers, allowing systematic training and evaluation of diverse expression patterns across nuclear, cytoplasmic, and membranous compartments. For machine learning purposes, we randomly split the dataset into a training set of 10,493,672 images and a held-out test set of 2,000 images.

\subsection*{Training an automated IHC evaluation AI model}
We then use the HPA10M dataset to train the iSight model using a multi-task learning framework that integrates visual features from IHC images with metadata to predict staining characteristics (\textbf{Figure \ref{Figure 1}d--e}). The model consists of three main components working together to analyze immunohistochemistry slides, namely, the tissue encoding module, the context encoding module, and the multi-task learning module.

The tissue encoding module processes whole slide images or tissue microarrays by dividing them into non-overlapping patches, each with 336×336 pixels in size. Each patch is independently processed through a Vision Transformer (openai/clip-vit-large-patch-14-336 \cite{radford2021learning}) to obtain a sequence of 577 token embeddings per patch (576 spatial tokens and one [CLS] token). All tokens are passed through a single linear projection. Rather than aggregating tokens within each patch, iSight concatenates all token embeddings across patches and uses gated attention to create a weighted tissue-level representation (\textbf{Figure \ref{Figure 1}d}).

Next, the context encoding module processes metadata associated with each sample. Cell type information is one-hot encoded and linearly projected to the latent space. During training, text captions are randomly introduced by concatenating SNOMED code, SNOMED text, marker gene, and cell type information (\textit{e.g.}, ``Pancreas. Pancreatic adenocarcinoma, NOS. Antibody (gene): CDK1 for tumor cells''). These captions are processed by the CLIP \cite{radford2021learning} text encoder and projected to the shared latent space. At inference time, the model processes only cell type information to avoid information leakage.

Finally, the multi-task learning framework \cite{zhang2018overview,zhang2021survey} uses parallel classification heads to simultaneously handle five prediction tasks (\textbf{Figure \ref{Figure 1}e}). The three primary tasks include staining intensity (4 classes: negative, weak, moderate, strong), staining location (4 classes: none, cytoplasmic/membranous, nuclear, cytoplasmic/membranous \& nuclear), and staining quantity (4 classes: none, <25\%, 25\%-75\%, >75\%). Two auxiliary tasks facilitate model learning: tissue type (58 classes) and malignancy status (2 classes: normal, cancer).
We reduced the 65 original tissue types to 58 classes by removing the "cancer" and "tissue" suffixes (e.g., "breast cancer" and "breast" were merged into "breast"), while retaining malignancy information through a separate prediction task.
Each head consists of a linear projection from the shared embedding space to task-specific outputs, enabling end-to-end optimization across all tasks simultaneously. iSight was trained end to end using the Adam optimizer with a learning rate of $1 \times 10^{-6}$ for one epoch on the HPA10M training set. Additional training details are provided in the \textbf{Methods} section.

\begin{figure}
    \centering
    \includegraphics[width=0.95\linewidth]{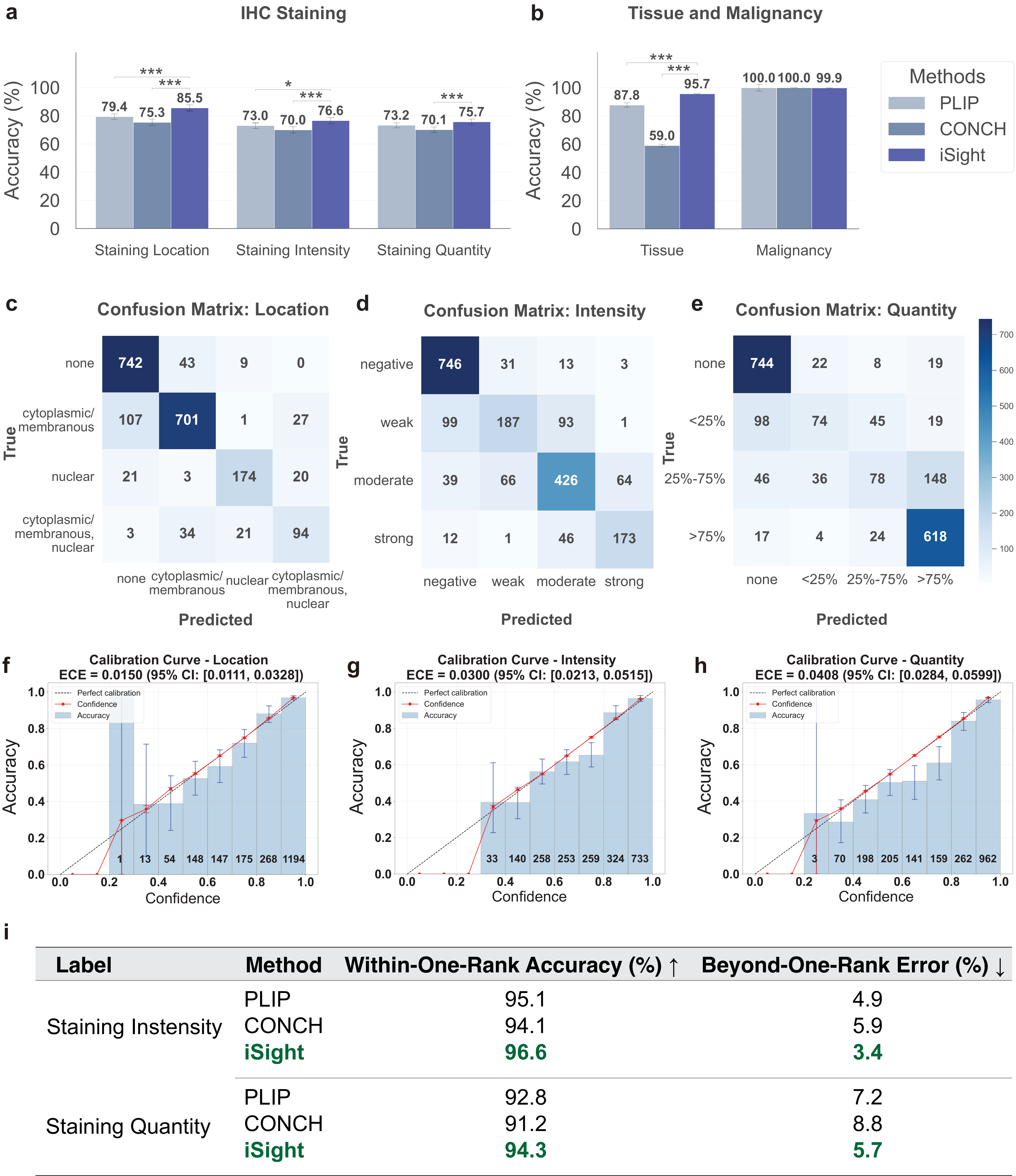}
    \caption{
    \textbf{Model comparison and performance analysis on held-out HPA test set.} 
\textbf{a}, Performance comparison between iSight, fine-tuned PLIP, and fine-tuned CONCH on three primary staining tasks: location, intensity, and quantity. 
\textbf{b}, Performance comparison on two auxiliary tasks: tissue type classification and malignancy detection. 
\textbf{c-e}, Confusion matrices for iSight showing classification patterns for staining intensity, location, and quantity respectively.
\textbf{f-h}, Calibration curves for iSight showing the relationship between predicted confidence and actual accuracy for staining location, intensity, and quantity tasks. Each curve displays Expected Calibration Error (ECE) values with 95\% confidence intervals. 
\textbf{i}, Within-one-rank accuracy and beyond-one-rank error rates for ordinal classification tasks (staining intensity and quantity) comparing iSight with baseline models PLIP and CONCH.
    Error bars represent 95\% confidence intervals derived from bootstrap resampling. Paired two-tailed Student’s t-tests were used to evaluate statistical significance (*p < 0.05, **p < 0.01, ***p < 0.001).
}
    \label{Figure 2}
\end{figure}

\subsection*{Analytical validation on the held-out HPA dataset}

After training iSight, we evaluated model performance on held-out datasets.
For comparison, we fully fine-tuned both baseline foundation models (PLIP and CONCH) on the same multi-task learning framework using the HPA10M training set. Because IHC scoring involves subjective, ordinal judgments with substantial inter-observer variability, even expert consensus labels do not constitute an absolute gold standard, making moderate accuracies still clinically meaningful. Nevertheless, iSight achieved strong performance across all three primary tasks on the held-out HPA test set of 2,000 images. Overall accuracies were 85.5\% for staining location, 76.6\% for staining intensity, and 75.7\% for staining quantity (\textbf{Figure \ref{Figure 2}a}). Accuracy on the two auxiliary tasks was 95.7\% for tissue type and 99.9\% for malignancy status (\textbf{Figure \ref{Figure 2}b}). 

When PLIP \cite{huang2023visual} and CONCH \cite{lu2024visual} were fine-tuned for the same classifications, both models underperformed on all primary tasks. For staining intensity, PLIP achieved 73.0\% (P-value=$ 2.35\times10^{-2 }$) accuracy and CONCH 70.0\% (P-value=$ 2.04\times10^{-5 }$), corresponding to absolute gaps of 3.6\% and 6.6\%. Comparable separations were observed for location (6.1\%, P-value=$ 6.01\times10^{-6 }$ and 10.3, P-value=$ 1.37\times10^{-13 }$) and quantity (2.5\%, P-value=$ 1.19\times10^{-1 }$ and 5.7\%, P-value=$ 4.34\times10^{-4 }$) (\textbf{Figure \ref{Figure 2}a}). For the auxiliary tasks (\textbf{Figure \ref{Figure 2}b}), iSight greatly outperforms the baselines for tissue classification, with PLIP achieving 87.8\% (P-value=$ 2.22\times10^{-16 }$) and CONCH 59.0\% (P-value $=0.00$), representing improvements of 7.9\% and 36.7\%, respectively.
However, all models perform comparably for malignancy prediction. This pattern suggests that our token-level attention pooling mechanism is most beneficial for tasks driven by fine-grained staining cues, whereas broader, image-level phenotypes such as malignancy gain less from this token-level weighting.

Analysis of the confusion matrices further demonstrated that iSight achieved strong performance across all three primary tasks (\textbf{Figure \ref{Figure 2}c--e}). For staining location, the model demonstrated excellent discrimination between different subcellular compartments, with particularly accurate classification of cytoplasmic/membranous versus nuclear staining patterns. For staining intensity and quantity, the model showed consistent performance with most prediction errors occurring between closely related categories.


Additionally, model calibration analysis also demonstrated that iSight produces well-calibrated predictions across all primary tasks (\textbf{Figure \ref{Figure 2}f-h}). The expected calibration error \cite{guo2017calibration} (ECE) values were consistently low: 0.0150 (95\% CI: [0.0111, 0.0328]) for staining location, 0.0300 (95\% CI: [0.0213, 0.0515]) for staining intensity, and 0.0408 (95\% CI: [0.0284, 0.0599]) for staining quantity. The calibration curves showed good alignment between predicted confidence and actual accuracy across different confidence bins, with staining location demonstrating the best calibration performance.

The within-one-rank accuracy and beyond-one-rank error metrics provided additional insights into model performance for ordinal tasks. For staining intensity, iSight achieved 96.6\% within-one-rank accuracy with only 3.4\% beyond-one-rank errors, substantially outperforming fine-tuned PLIP (95.1\% and 4.9\%) and CONCH (94.1\% and 5.9\%). Similar patterns were observed for staining quantity, where iSight achieved 94.3\% within-one-rank accuracy and 5.7\% beyond-one-rank errors, compared to fine-tuned PLIP (92.8\% and 7.2\%) and CONCH (91.2\% and 8.8\%) (\textbf{Figure \ref{Figure 2}i}). For each task, the baselines showed lower within-one-rank accuracies and larger severe miscalculation errors.

\subsection*{iSight maintains robust performance under image quality variations}

To evaluate iSight's robustness to common image quality variations encountered in clinical practice, we further assessed model performance under synthetic image corruptions at four severity levels (\textbf{Supplementary Figure \ref{Supple Figure 2}}). We applied two types of perturbations to the HPA test set: random cutout and salt-and-pepper noise. Random cutout simulates tissue folding, scanning artifacts, or information loss by masking random rectangular regions. Salt-and-pepper noise models pixel-level corruption during image acquisition.

Across all perturbation types and severity levels, iSight maintained stable performance with minimal degradation.
Under random cutout noise, iSight demonstrated robust performance across all three tasks. For staining location (\textbf{Supplementary Figure \ref{Supple Figure 2}a}), accuracy remained between 0.853-0.856 across all severity levels compared to the baseline of 0.856, with a maximum drop of only 0.3\%. For staining intensity (\textbf{Supplementary Figure \ref{Supple Figure 2}b}), accuracy ranged from 0.757-0.762 versus baseline 0.765. For staining quantity (\textbf{Supplementary Figure \ref{Supple Figure 2}c}), accuracy remained between 0.750-0.759 compared to baseline 0.757.
Under salt-and-pepper noise, the model maintained similarly stable performance. For staining location (\textbf{Supplementary Figure \ref{Supple Figure 2}d}), accuracy ranged from 0.841-0.860, with a maximum deviation of 1.5\% from baseline. For staining intensity (\textbf{Supplementary Figure \ref{Supple Figure 2}e}), accuracy varied from 0.753-0.763. For staining quantity (\textbf{Supplementary Figure \ref{Supple Figure 2}f}), accuracy ranged from 0.744-0.763.
These results demonstrate that iSight's predictions are resilient to common image quality variations encountered in clinical practice, including both structured occlusions and random pixel-level corruption.

\subsection*{Evaluating the effect of AI assistance on pathologist IHC inference}

\begin{figure}
    \centering
    \includegraphics[width=1.0\linewidth]{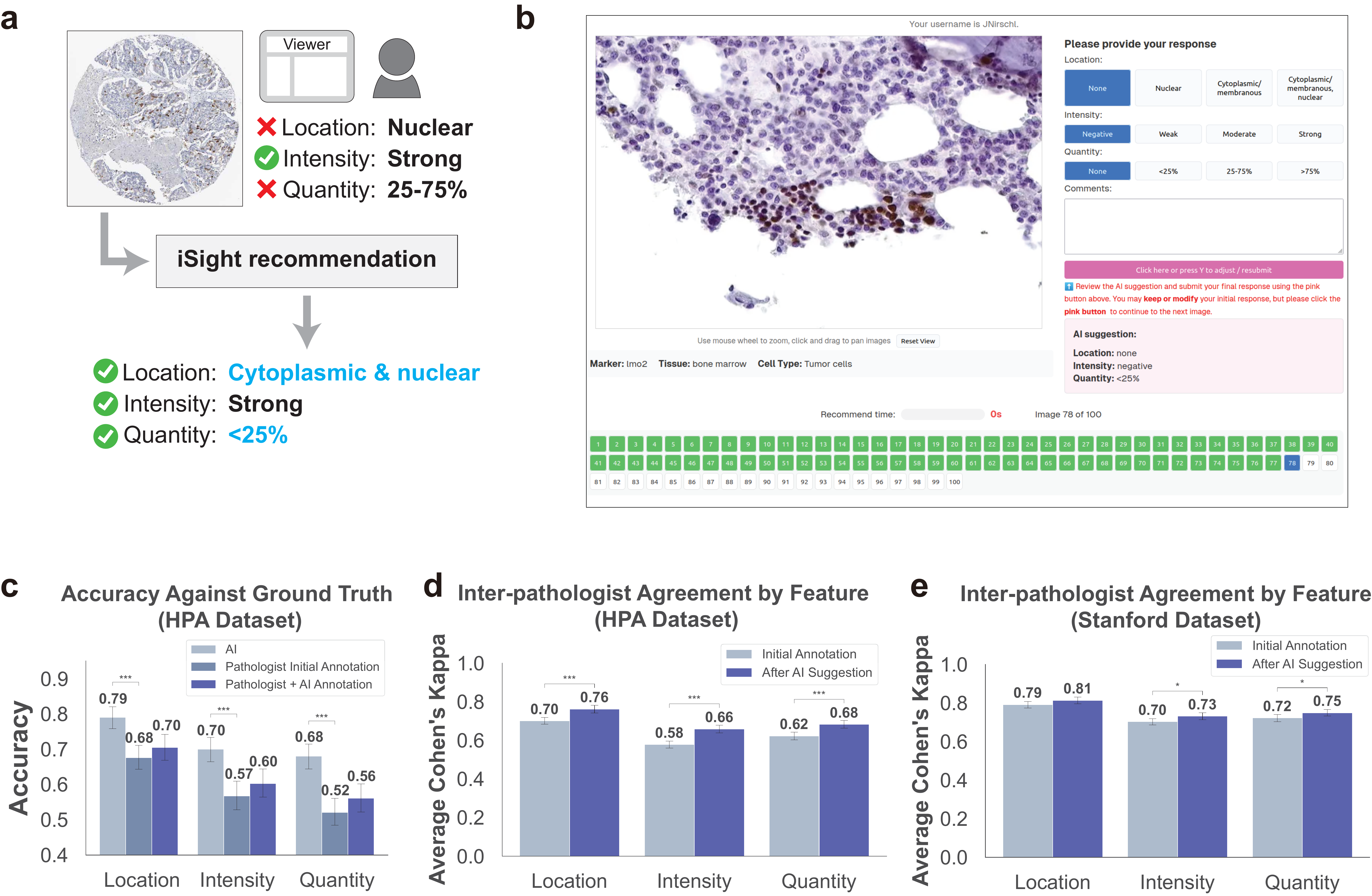}
    \caption{\textbf{Pathologist user study across HPA and Stanford datasets.} 
    \textbf{a}, iSight pathologist user study workflow. 
    \textbf{b}, Web-based annotation interface for pathologist user study.
    \textbf{c}, Comparison of classification performance against ground truth in the HPA dataset for AI predictions, pathologist initial annotations, and pathologist annotations after AI suggestion across three staining tasks. 
    \textbf{d-e}, Inter-pathologist agreement measured by average Cohen's $\kappa$ before and after AI suggestions for each feature in HPA and Stanford datasets.
    Error bars represent 95\% confidence intervals derived from bootstrap resampling. Paired two-tailed Student’s t-tests were used to evaluate statistical significance (*p < 0.05, **p < 0.01, ***p < 0.001).
}
    \label{Figure 3}
\end{figure}

To evaluate the clinical utility of iSight, we conducted a user study with eight pathologists (three first-year residents, one second-year resident, one fellow, and three attending pathologists) (\textbf{Figure \ref{Figure 3}a}). Each pathologist independently annotated 200 images (100 from HPA and 100 from Stanford TMAD) for staining location, intensity, and quantity. The annotation process consisted of two phases: first, pathologists provided their initial annotations based on the tissue image, marker gene, expected localization, tissue type, and cell type, without iSight AI assistance; second, pathologists were shown the iSight prediction and given the option to revise their answers or maintain their original judgment. 
We developed a web-based annotation interface to facilitate this process (\textbf{Figure \ref{Figure 3}b}). The interface displayed the tissue image with zoom functionality, metadata, and AI recommendations side-by-side, allowing pathologists to systematically evaluate and optionally adopt the model's suggestions.

In the HPA dataset, when comparing against ground truth labels, initial pathologist accuracies were 68\% for location, 57\% for intensity, and 52\% for quantity. After receiving AI suggestions, accuracies improved to 70\%, 60\%, and 56\%, respectively (\textbf{Figure \ref{Figure 3}c}). In comparison, iSight achieved accuracies of 79\%, 70\%, and 68\% for location, intensity, and quantity, which demonstrates superior performance to both initial and AI-assisted pathologist annotations.
Prior to AI assistance, pathologists showed an average inter-rater agreement of 0.63 across all tasks (Cohen's $\kappa$: 0.70 for location, 0.58 for intensity, 0.62 for quantity). After viewing iSight model predictions and being allowed to adjust initial annotations, the inter-rater agreement increased to 0.70 (Cohen's $\kappa$: 0.76 for location, 0.66 for intensity, 0.68 for quantity) (\textbf{Figure \ref{Figure 3}d}).

For the Stanford TMAD dataset, ground-truth labels are unavailable; therefore, accuracy metrics are not reported. Instead, we evaluated inter-rater agreement and patterns of AI influence. The initial inter-rater agreement was 0.74 across all tasks (Cohen's $\kappa$: 0.79 for location, 0.70 for intensity, 0.72 for quantity). After AI suggestion, agreement increased modestly to 0.76 (Cohen's $\kappa$: 0.81 for location, 0.73 for intensity, 0.75 for quantity) (\textbf{Figure \ref{Figure 3}e}), with statistically significant improvements observed for intensity and quantity.

\subsection*{Quantifying the influence of AI assistance on pathologist decision-making}

\begin{figure}
    \centering
    \includegraphics[width=1.0\linewidth]{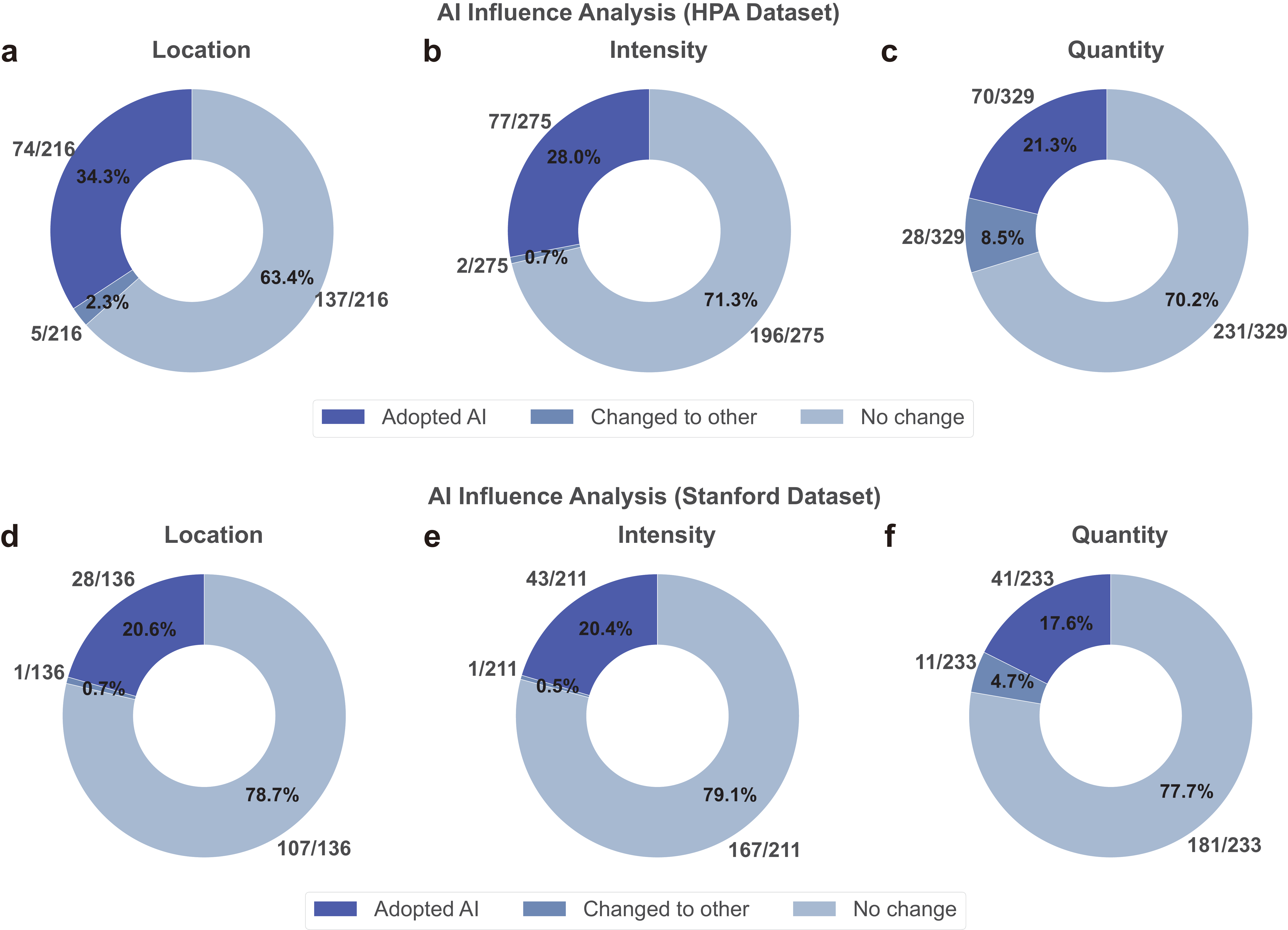}
    \caption{\textbf{AI influence analysis across HPA and Stanford datasets.} 
    \textbf{a-c}, AI influence analysis in the HPA dataset, which shows the distribution of pathologist responses when their initial annotation disagreed with AI predictions for location, intensity, and quantity tasks respectively: adopted AI suggestion (dark blue), changed to alternative label (medium blue), or maintained original annotation (light blue).
    \textbf{d-f}, AI influence analysis in the Stanford dataset for location, intensity, and quantity tasks.
}
    \label{Figure 4}
\end{figure}

We analyzed cases where pathologists' initial annotations disagreed with iSight predictions to understand how AI influences clinical decision-making. In the HPA dataset, there were 216 location disagreements, 275 intensity disagreements, and 329 quantity disagreements between initial pathologist annotations and AI predictions.

For location, pathologists adopted the AI suggestion in 74 cases (34.3\%), kept their original judgment in 137 cases (63.4\%), and changed to a different label in 5 cases (2.3\%) (\textbf{Figure \ref{Figure 4}a}). For intensity, 77 cases (28.0\%) adopted AI, 196 cases (71.3\%) remained unchanged, and 2 cases (0.7\%) chose alternatives (\textbf{Figure \ref{Figure 4}b}). For quantity, 70 cases (21.3\%) adopted AI, 231 cases (70.2\%) remained unchanged, and 28 cases (8.5\%) chose alternatives (\textbf{Figure \ref{Figure 4}c}). Among the three tasks, location showed the highest AI adoption rate at 34.3\%, followed by intensity at 28.0\% and quantity at 21.3\%.

We observed similar trends in the Stanford TMAD dataset. For location, 28 out of 136 disagreements (20.6\%) adopted AI, 107 cases (78.7\%) remained unchanged, and 1 case (0.7\%) chose an alternative (\textbf{Figure \ref{Figure 4}d}). For intensity, 43 out of 211 disagreements (20.4\%) adopted AI, 167 cases (79.1\%) remained unchanged, and 1 case (0.5\%) chose an alternative (\textbf{Figure \ref{Figure 4}e}). For quantity, 41 out of 233 disagreements (17.6\%) adopted AI, 181 cases (77.7\%) remained unchanged, and 11 cases (4.7\%) chose alternatives (\textbf{Figure \ref{Figure 4}f}). Similarly, location demonstrated the highest AI adoption rate at 20.6\%, followed by intensity at 20.4\% and quantity at 17.6\%. These findings demonstrate that AI predictions can effectively influence pathologist decision-making and assist in making better diagnostic judgments. Both experiments demonstrate the value of human-AI collaboration, where iSight serves as a decision support tool that pathologists can critically evaluate and integrate into their clinical reasoning process.

\section*{Discussion}

In this work, we introduce iSight, a multi-task learning framework for automated immunohistochemistry (IHC) staining assessment. By jointly modeling staining intensity, subcellular localization, staining quantity, and tissue context, iSight directly addresses the long-standing challenges of subjectivity, limited reproducibility, and increasing workload in IHC interpretation. Given the central role of IHC in cancer diagnosis, prognostic stratification, and therapeutic decision-making, this capability has the potential to improve diagnostic consistency at scale. One of the major contribution in this work is the curation of the HPA10M dataset from over two decades of data collection in Human Protein Atlas. The HPA10M dataset provides comprehensive IHC coverage with 10,495,672 IHC images across 58 tissue types. Unlike H\&E-focused datasets, this is the first large collection of IHC images, enables systematic evaluation of IHC-specific patterns. The dataset's scale represents a significant advance over existing pathology datasets, encompassing over 17,200 protein targets across 45 normal tissue types and 20 cancer types.

In our evaluation, the in-house model ``iSight'' achieved 76.6\% accuracy for intensity, 85.5\% for location, and 75.7\% for quantity on held-out data, outperforming fine-tuned pathology foundation models by 2.5-10.2\%. These results demonstrate that task-specific architectural innovations can provide meaningful improvements over generic pathology foundation models. Beyond accuracy, the within-one-rank accuracy of 96.6\% for intensity and 94.3\% for quantity indicates that errors are typically minor disagreements rather than severe misclassifications. Low beyond-one-rank error rates (3.4\% for intensity, 5.7\% for quantity) suggest iSight rarely makes egregious mistakes. The expected calibration errors (0.0150--0.0408) also indicate that confidence scores accurately reflect true accuracy, which enables clinicians to appropriately weight AI predictions.

Beyond the dataset contribution and model development, this study is significant in demonstrating how AI can be integrated into real-world pathology workflows as a decision support system rather than a replacement for expert judgment. Through a controlled pathologist reader study spanning two independent datasets, we show that AI assistance can improve diagnostic accuracy and inter-observer agreement while preserving pathologist autonomy. In our user study, eight pathologists evaluated 200 images, with inter-rater agreement improving modestly after AI assistance. On HPA images with ground truth, iSight outperformed pathologists' initial annotations (79\% vs 68\% for location, 70\% vs 57\% for intensity, 68\% vs 52\% for quantity). These results suggest that pathologists can experience large discordance on IHC evaluation and demonstrate that AI can, on average, achieve performance exceeding expert assessments. Importantly, pathologist accuracy improved after expert viewing AI suggestions, demonstrating that human-AI collaboration enhances performance beyond either alone. Interestingly, although pathologist performance improved after reviewing AI suggestions, it remained lower than that of the AI model alone. This gap likely reflects cognitive and psychological factors inherent to expert decision-making, including anchoring to initial judgments, loss aversion when revising decisions, and a natural reluctance to fully override one’s own interpretation. These observations motivate future studies focused on understanding pathologist behavior and decision dynamics in human–AI interaction, with the goal of designing interfaces and feedback mechanisms that better support effective and confident adoption of AI guidance. Meanwhile, the observed improvement after pathologists incorporated AI suggestions highlights the practical value of AI assistance and motivates future deployment of such systems in routine clinical and research settings.

Several limitations still exist in this study. First, external validation is limited. While we evaluated on Stanford TMAD samples, multi-institutional studies across different scanners and staining protocols are needed to assess generalizability. Second, the dataset includes only images and text. Incorporating additional modalities such as H\&E images, laboratory results and patient history could improve diagnostic accuracy, while beyond our current scope, but could be an interesting future research. In clinical practice, IHC is interpreted with morphological assessment, clinical history, and molecular results. Future iterations could incorporate paired H\&E and IHC images and clinical metadata through multimodal learning architectures. Third, the model's performance on rare staining patterns remains uncertain. While the training dataset is large, unusual expression patterns, rare tumor types, and technically challenging cases may be underrepresented. Lastly, the reference labels provided by the Human Protein Atlas may not represent an absolute gold standard ground truth for IHC interpretation; therefore, it is not surprising that neither iSight nor other fine-tuned pathology foundation models achieve accuracies exceeding 90\% for staining location, intensity, and quantity. Looking forward, this limitation underscores the need for more rigorous and diverse reference standards for IHC evaluation. Future work may need to incorporate multi-expert consensus annotations, and probabilistic or soft labels that capture inter-observer variability.

Despite these limitations, this work represents the first large-scale, systematically curated immunohistochemistry dataset and the first comprehensive study to demonstrate the value of AI in assisting, rather than replacing, pathologists in IHC interpretation. HPA10M enables foundation-scale learning of IHC-specific staining patterns, while iSight provides calibrated, efficient, and robust predictions across multiple tasks. Future work should expand rare disease coverage, incorporate multi-modal clinical data, conduct prospective clinical studies, and develop enhanced explainability features. As IHC volume continues to expand, such human–AI collaborative systems will be essential for reducing variability, improving consistency, and scaling pathology workflows.

\clearpage
\renewcommand{\figurename}{Supplementary Figure}  
\setcounter{figure}{0}   

\section*{Methods}

\subsection*{Dataset overview}
\subsubsection*{Human Protein Atlas}

The HPA10M dataset comprises 10,495,672 immunohistochemistry (IHC) images derived from version 23.0 of the Human Protein Atlas (HPA) \footnote{https://www.proteinatlas.org/}, released on June 19, 2023. This release includes antibody-based protein profiling for over 17,200 human genes (87\% of the human proteome), with annotations spanning normal and cancerous tissues. Each image in HPA10M is associated with detailed metadata, including tissue source, SNOMED-coded diagnosis, patient age and sex, and molecular targets indexed via UniProt and ENSEMBL identifiers.

Dataset construction followed a multi-stage curation and quality control pipeline. Raw HPA image and metadata archives were parsed and harmonized using standardized biomedical ontologies, including Uberon for anatomy and SNOMED CT for diagnoses. Systematic errors and inconsistencies were corrected, and redundant image replicates were downsampled to ensure balanced class representation. Only images with complete, high-confidence metadata were retained.
The curated dataset encompasses 45 normal tissue types and 20 major cancer types. This comprehensive coverage of both normal and pathological tissues enables the model to learn discriminative features across the full spectrum of tissue morphologies and disease states.

To enable large-scale training of multimodal AI models, the curated data were serialized into WebDataset format, with each image paired to its corresponding JSON-formatted metadata. Descriptive captions summarizing staining intensity, localization, and quantity were generated to support vision-language learning tasks. Additional demographic, phenotypic, and molecular annotations are preserved in tabular form using Feather format for efficient querying and downstream integration. The dataset encompasses a diverse range of histopathological diagnoses (\textbf{Supplementary Figure \ref{Supple Figure 3}}), with predominant categories including normal tissue NOS, adenocarcinoma NOS, squamous cell carcinoma NOS and various other carcinoma subtypes.

This richly annotated dataset provides a scalable foundation for developing and evaluating AI models for IHC interpretation. All images are traceable via cryptographic MD5 checksums, and links to original HPA subject identifiers facilitate integration with additional omics and clinical endpoints.

\subsubsection*{Stanford Tissue Microarray Database}
The Stanford Tissue Microarray Database (TMAD) \cite{marinelli2007stanford} \footnote{https://tma.im/cgi-bin/home.pl} is a public repository of tissue microarray images and metadata, providing tools for image annotation, scoring, and analysis. It contains over 200,000 images from chromogenic and fluorescence-stained tissue microarrays, covering a broad range of disease tissues and biomarkers. For external validation purposes, we curated 100 IHC images from the database, selecting clinically relevant biomarkers with a mix of nuclear and cytoplasmic staining patterns to supplement the pathologist user study.

\subsection*{Model architecture and training}
We developed a multi-class, weakly supervised multiple-instance learning model for tissue-level classification. The model combines a pretrained Vision Transformer (ViT) for patch-level feature extraction with a gated attention mechanism to aggregate token-level feature information across tissue patches. The model is designed to handle heterogeneous tissue structures by enabling fine-grained spatial attention over sub-patch regions. Additionally, iSight incorporates metadata such as cell type, marker, and textual descriptions of the tissue in a context encoding module. 

\subsubsection*{Tissue encoding module}
Each tissue image is divided into a set of non-overlapping patches of size $336\times336$ pixels. The patches are independently processed through a ViT (openai/clip-vit-large-patch-14-336 \cite{radford2021learning}) to obtain a sequence of 577 token embeddings per patch, 576 spatial tokens and one [CLS] token. All tokens are then passed through a single linear projection. 

Rather than aggregating tokens within each patch, iSight performs attention-based \cite{vaswani2017attention} pooling over the entire sequence of tokens across all patches in a tissue image. Specifically, all token embeddings from all patches are concatenated into a single sequence, and a gated attention network \cite{ilse2018attention} assigns a learned importance weight to each token. The attention weights are normalized via a softmax function across the entire set of tokens and used to create a weighted sum of the token embeddings, resulting in a single tissue-level representation.

\subsubsection*{Context encoding module}
In addition to the tissue image, iSight processes metadata associated with each sample. Specifically, the cell type is one-hot encoded and linearly projected to the latent space of the tissue encoding module. Lastly, during training we randomly introduce text captions by concatenating the SNOMED code, SNOMED text, marker gene, and cell type. The caption is processed by the text encoder of the CLIP model (openai/clip-vit-large-patch-14-336) and passed through a linear projection to the shared latent space. During inference, iSight only processes the cell type, not the text captions.

\subsubsection*{Multi-task classification heads}
Finally, the tissue and context embeddings are aggregated and passed to a set of parallel classification heads, each corresponding to a distinct prediction task. The three target tasks of interest include staining intensity (four classes–negative, weak, moderate, strong), staining location (four classes–none, cytoplasmic/membranous, nuclear, cytoplasmic/membranous and nuclear), staining quantity (four classes–none, <25\%, 25\% - 75\%, >75\%). In addition to the target tasks, we include two auxiliary tasks to facilitate model learning–tissue type (58 tissue classes), and malignancy status (two classes–normal, cancer). Each head consists of a linear projection from the shared embedding space to the task-specific output dimensionality.

\subsubsection*{Optimization and training hyperparameters}
The model is optimized end-to-end via Adam \cite{kingma2014adam} over the multi-class cross-entropy loss for each prediction with equal weighting for each task. We used a learning rate of 1e-6 and a weight decay of 1e-5, and trained for one epoch. The model was trained on two NVIDIA H100 SXM5 80GB GPUs with batch size = 2.

\subsection*{Model validation}
\subsubsection*{Analytical validation}
We validated iSight on a held-out test set of 2,000 images from the HPA dataset, stratified to match the class distribution of the full HPA dataset. Class-wise accuracy of the target tasks was reported along with 95\% confidence intervals estimated via bootstrap resampling (n=1,000). 

In addition to empirical accuracy, we introduce two further performance metrics, ``within-one-rank accuracy'' and ``beyond-one-rank error'' for staining intensity and staining quantity. Due to the inter-rater variability in labeling and the fact that these labels are ordinal, minor deviations between predicted and ground truth classes (\textit{e.g.}, predicting ``moderate'' instead of ``strong'' staining) may be less clinically significant than larger errors. The within-one-rank accuracy captures predictions that fall within ±1 class of the ground truth. In contrast, the beyond-one-rank error captures more substantial misclassifications, providing a more nuanced assessment of model performance in the context of ordinal outcomes.

To assess model calibration across the prediction tasks, we calculated the expected calibration error (ECE) \cite{guo2017calibration} by comparing predicted confidence scores with the empirical accuracies in a bin-wise manner. Within each task, class probabilities were computed by the softmax of the raw logits, and the predicted confidence was defined as the maximum class probability. We discretized the confidence range [0, 1] into 10 equally spaced bins, and computed the absolute difference between average confidence and accuracy within each bin. The final ECE was the weighted average of the bin-wise differences, normalized by the number of samples. To quantify uncertainty, we performed bootstrap resampling (n=1,000) to calculate 95\% confidence intervals for ECE and per-bin metrics.

\subsubsection*{Baseline comparison}
We compare the performance of iSight against two histopathology foundation models. This includes PLIP \cite{huang2023visual}, a vision-language model pretrained on a large multimodal dataset of over 200,000 histopathology images paired with natural-language descriptions curated from publicly shared medical content, and CONCH \cite{lu2024visual}, a vision-language model pretrained on over 1.17 million paired histopathology images and captions, derived from public sources. Each model was fully fine-tuned for the multi-task classification task using the HPA training dataset by attaching classification heads to the base model vision encoder projections. Images were resized to the correct dimensions for each model.

\subsubsection*{Pathologist user study}
We conducted a user study to evaluate the alignment of iSight with expert annotations and its potential impact on human decision-making. Eight pathologists (three first-year residents, one second-year resident, one fellow, and three attending pathologists) independently annotated 200 images, 100 from the held-out HPA dataset and 100 from the external Stanford TMA dataset. The pathologists were asked to annotate the staining location, intensity, and quantity based on the tissue image, the marker gene, the expected localization, the tissue type, and the cell type. Each annotation is conducted in two phases. First each pathologist was asked to provide their initial annotation. Second, the pathologists were shown the iSight prediction (AI suggestion) and given the option to revise their answers or leave them unchanged.

For this experiment, we evaluated four metrics. First, two analytical metrics, accuracy against ground truth labels from the HPA dataset and inter-rater agreement over all images, before and after the AI suggestion. Inter-rater agreement was computed as the mean pairwise agreement: for each image we enumerated every unique pair of the eight pathologists (up to 28 pairs), assigned a score of 1 when the two labels matched and 0 otherwise, and averaged these 0/1 values across all images, yielding a value from 0 (no concordance) to 1 (perfect concordance). Second, we measured the impact of the AI suggestion on the user annotation by examining AI influence and adoption. AI influence was defined as the proportion of cases in which the pathologist’s initial annotation disagreed with the iSight prediction, and the pathologist subsequently changed their answer after viewing the model output. Additionally, we further categorized these cases based on whether the pathologist left their annotation unchanged, revised it to an alternative label, or adopted the AI suggestion as their final decision.

\subsubsection*{Model robustness validation}

We evaluated model robustness using two types of synthetic image corruptions applied to the HPA test set at four severity levels (\textbf{Supplementary Figure \ref{Supple Figure 2}}).
Salt-and-pepper noise randomly replaces pixels with either white (salt) or black (pepper) values with equal probability. Severity levels were defined by the percentage of affected pixels: Level 1 (1\% total: 0.5\% salt + 0.5\% pepper), Level 2 (2\%: 1\% + 1\%), Level 3 (5\%: 2.5\% + 2.5\%), and Level 4 (8\%: 4\% + 4\%). The noise was applied uniformly across the entire image, simulating pixel-level corruption during image acquisition or compression.
Random cutout masks rectangular regions to simulate tissue folding, scanning artifacts, or missing information. Each severity level applies multiple independent cutout operations: Level 1 (1 cutout, maximum 5\% of image area), Level 2 (2 cutouts, 8\% each), Level 3 (3 cutouts, 10\% each), and Level 4 (4 cutouts, 15\% each). For each cutout operation, the actual area is randomly sampled between 2\% and the specified maximum area, the aspect ratio is randomly selected between 0.3 and 3.0 to create realistic rectangular occlusions, and cutouts are filled with either black (intensity = 0) or gray (intensity = 0.5) randomly selected. Cutout positions were fully randomized across the image.

\section*{Acknowledgements}

This project was supported by the startup funding from the Perelman School of Medicine, University of Pennsylvania.

\section*{Ethics declarations}
The authors declare no competing interests.

\section*{Author contributions statement}

Jacob Leiby, Jialu Yao, Pan Lu, James Zou, Thomas Montine, Jeffrey Nirschl, Zhi Huang conceived the experiments. Jacob Leiby, Jialu Yao, Pan Lu, Jeffrey Nirschl, Zhi Huang conducted the experiments. Jialu Yao and Jacob Leiby analyzed the results. George Hu and Jeffrey Nirschl collected the dataset. Anna Davidian, Shunsuke Koga, Olivia Leung, Pravin Patel, Isabella Tondi Resta, Rebecca Rojansky, Derek Sung, Eric Yang participated in the pathologist user study experiments. Paul J. Zhang, Emma Lundberg, Dokyoon Kim, Serena Yeung-Levy provided valuable insights on immunohistochemistry study. James Zou, Thomas Montine, Jeffrey Nirschl, Zhi Huang supervised the project. All authors reviewed the manuscript.

\section*{Data Availability}

HPA10M is a curated dataset derived from the Human Protein Atlas and distributed under the Creative Commons Attribution-ShareAlike 4.0 International License, available at \href{https://huggingface.co/datasets/nirschl-lab/hpa10m}{https://huggingface.co/datasets/nirschl-lab/hpa10m}.

\section*{Code Availability}

Source code of iSight is available at \href{https://github.com/zhihuanglab/iSight}{https://github.com/zhihuanglab/iSight}.

\bibliography{references}

\clearpage
\label{MainTextLastPage}
\renewcommand{\MainLastPage}{MainTextLastPage}
\clearpage
\appendix

\setcounter{page}{1}

\renewcommand{\figurename}{Supplementary Figure}
\renewcommand{\tablename}{Supplementary Table}
\renewcommand{\thefigure}{S\arabic{figure}}
\renewcommand{\thetable}{S\arabic{table}}
\setcounter{figure}{0}
\setcounter{table}{0}

\fancyhf{}                      
\rfoot{\small\sffamily\bfseries\thepage}
\renewcommand{\headrulewidth}{0pt}
\renewcommand{\footrulewidth}{0pt}

\pagestyle{fancy}
\thispagestyle{fancy}

\input{supplementary_information}

\end{document}

%% file: supplementary_information.tex
\renewcommand{\rmdefault}{phv}   
\renewcommand{\sfdefault}{phv}   
\renewcommand{\familydefault}{\sfdefault}
\setlength{\parindent}{0pt}      

{\LARGE\bfseries Supplementary Information for}\\[6pt]
{\LARGE\bfseries iSight: Towards AI-assisted Automatic Immunohistochemistry Staining Assessment}

\vspace{2em}

\renewcommand{\figurename}{Supplementary Figure}  
\renewcommand{\tablename}{Supplementary Table}  
\setcounter{figure}{0}
\setcounter{table}{0}

\section*{Supplementary Figures}

\begin{figure}[h!]
    \centering
    \includegraphics[width=1.0\linewidth]{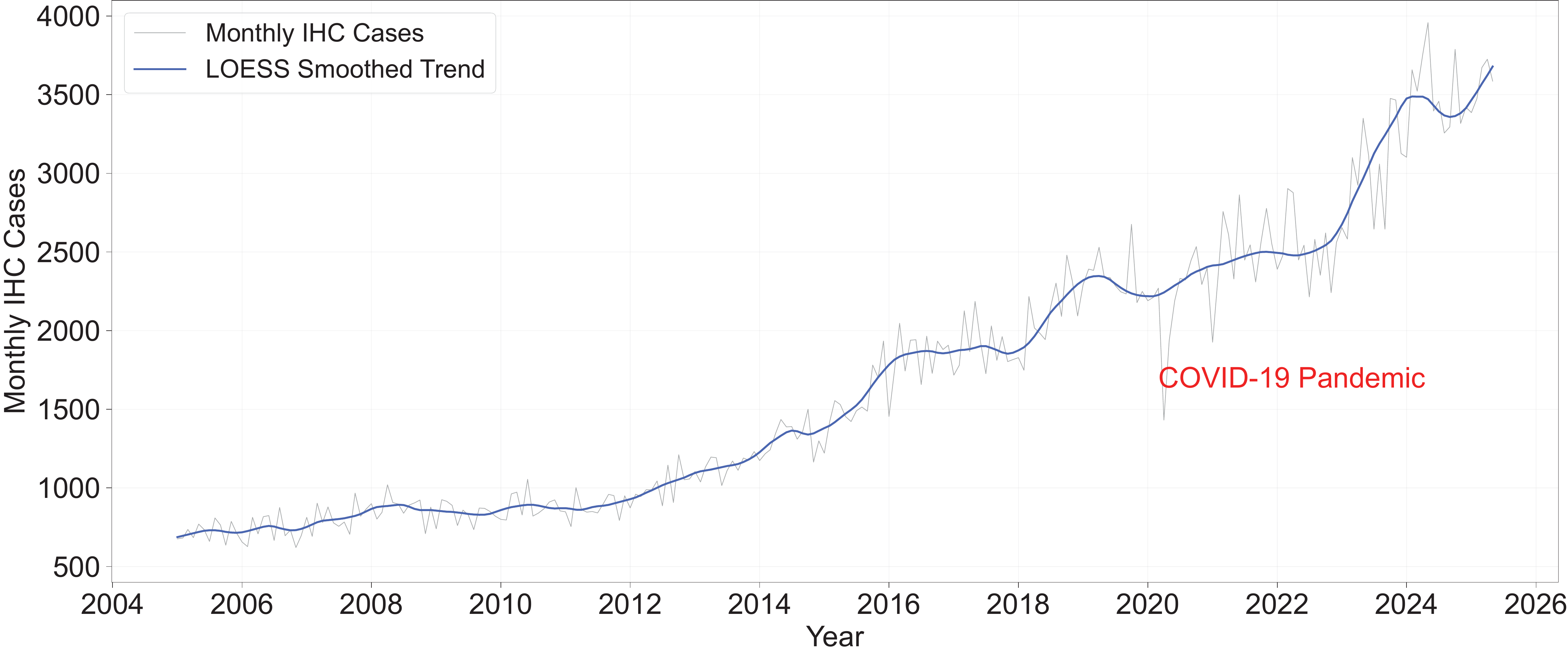}
    \caption{\textbf{Monthly IHC Case Volume at Stanford Healthcare from 2005 to 2025.} The gray line represents the raw monthly IHC case counts, while the blue line shows the LOESS smoothed trend. The data demonstrates a consistent upward trajectory in IHC utilization over the 20-year period, with notable fluctuations during the COVID-19 pandemic (marked in red text) starting in April 2020. The analysis includes 245 months of data, showing an increase from approximately 700 cases per month in 2005 to over 3,500 cases per month by 2025.
}
    \label{Supple Figure 1}
\end{figure}

\begin{figure}[h!]
    \centering
    \includegraphics[width=1.0\linewidth]{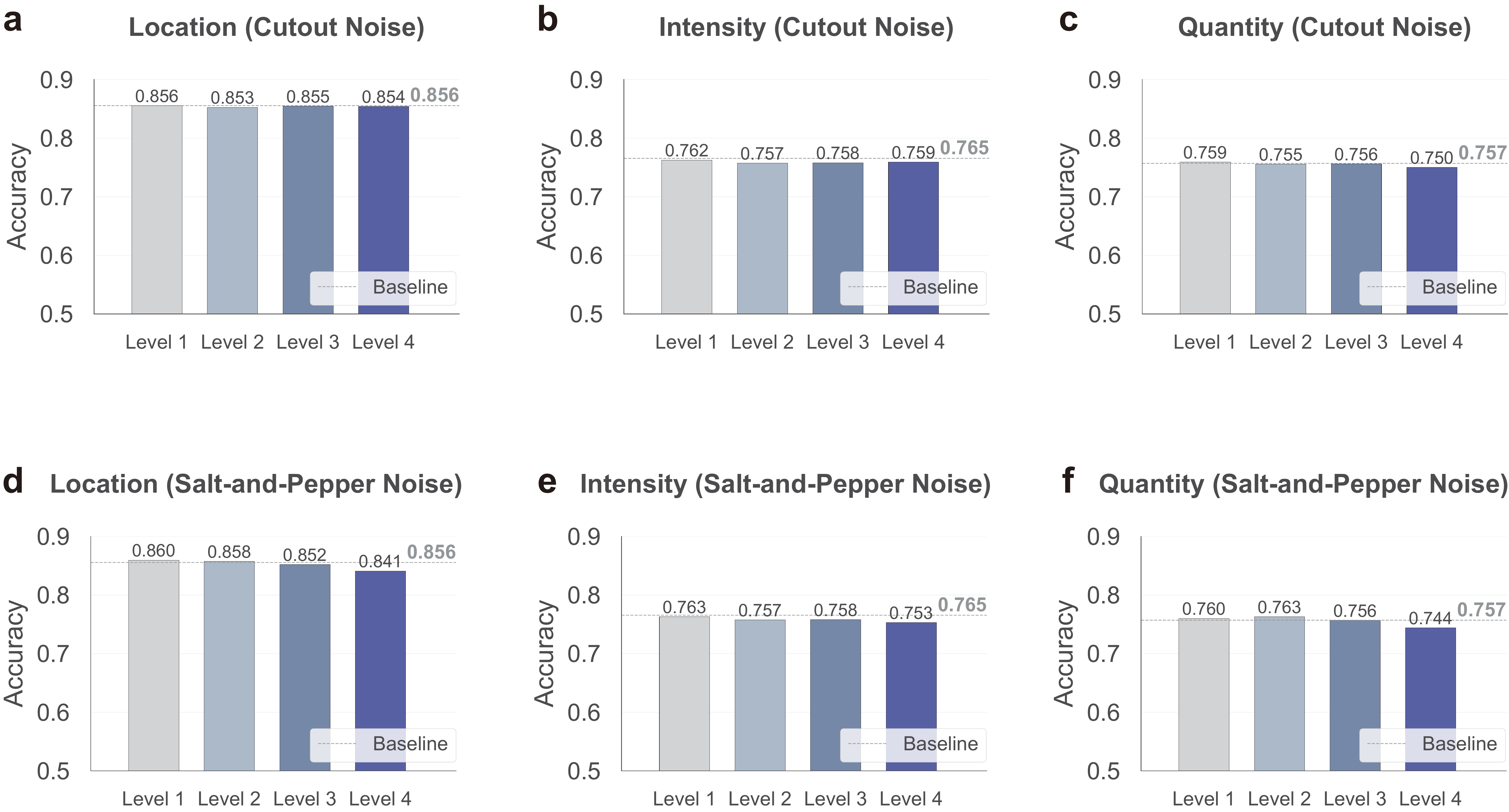}
    \caption{\textbf{Model robustness evaluation on HPA test set.} 
    \textbf{a-c}, Model accuracy on the HPA test set under random cutout noise at four severity levels for location (a), intensity (b), and quantity (c) prediction tasks. 
    \textbf{d-f}, Model accuracy under salt-and-pepper noise at four severity levels for location (d), intensity (e), and quantity (f) tasks. Baseline (dashed line) represents performance on unperturbed images. 
}
    \label{Supple Figure 2}
\end{figure}

\begin{figure}[h!]
    \centering
    \includegraphics[width=0.7\linewidth]{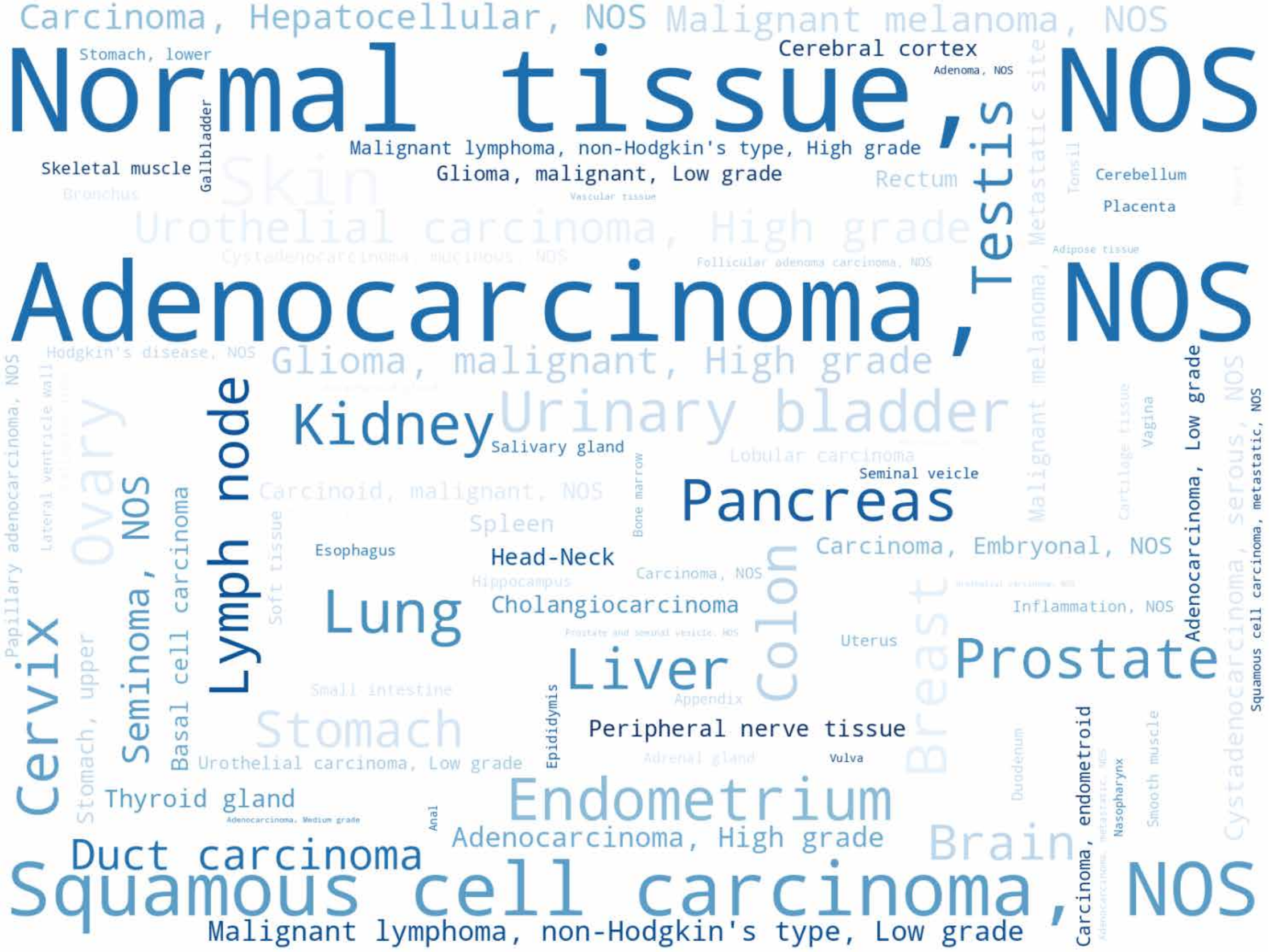}
    \caption{\textbf{SNOMED diagnosis term distribution in the HPA10M dataset.} Word cloud visualization shows the relative frequency of diagnostic terms, with text size proportional to occurrence frequency.
}
    \label{Supple Figure 3}
\end{figure}